\documentclass[sigconf]{acmart}

\renewcommand\footnotetextcopyrightpermission[1]{} % removes footnote with conference information in first column
\AtBeginDocument{%
  \providecommand\BibTeX{{%
    \normalfont B\kern-0.5em{\scshape i\kern-0.25em b}\kern-0.8em\TeX}}}

\setcopyright{none} 

\begin{document}

\title{Customizing Knowledge Graph Embedding to Improve Clinical Study Recommendation\vspace{0.2cm}}

\author{Xiong Liu}
\affiliation{%
  \institution{AI Innovation Lab, Novartis}
  \streetaddress{1 Th{\o}rv{\"a}ld Circle}
  \city{Cambridge}
  \country{USA}}

\author{Iya Khalil}
\affiliation{%
  \institution{AI Innovation Lab, Novartis}
  \city{Cambridge}
  \country{USA}
}

\author{Murthy Devarakonda}
\affiliation{%
  \institution{AI Innovation Lab, Novartis}
  \city{Cambridge}
  \country{USA}
  \vspace{0.5cm}% Additional space
}

\begin{abstract}
Inferring knowledge from clinical trials using knowledge graph embedding is an emerging area. However, customizing graph embeddings for different use cases remains a significant challenge. We propose custom2vec, an algorithmic framework to customize graph embeddings by incorporating user preferences in training the embeddings. It captures user preferences by adding custom nodes and links derived from manually vetted results of a separate information retrieval method. We propose a joint learning objective to preserve the original network structure while incorporating the user’s custom annotations. We hypothesize that the custom training improves user-expected predictions, for example, in link prediction tasks. We demonstrate the effectiveness of custom2vec for clinical trials related to non-small cell lung cancer (NSCLC) with two customization scenarios: recommending immuno-oncology trials evaluating PD-1 inhibitors and exploring similar trials that compare new therapies with a standard of care. The results show that custom2vec training achieves better performance than the conventional training methods. Our approach is a novel way to customize knowledge graph embeddings and enable more accurate recommendations and predictions.
\end{abstract}


\maketitle
% remove running header
\pagestyle{plain}

\section{Introduction}

Clinical trial design is a complex process, which involves searching relevant trials \cite{rybinski2020clinical}, extracting entities and relations \cite{liu2021clinical}, and comparing similar studies \cite{koroleva2019measuring}. To accelerate clinical development, innovative methods are needed to infer knowledge from existing clinical trials for new trial design.

Identifying clinical trials relevant to a specific user query is a classical NLP problem and can be challenging due to the complexity of clinical trials. ClinicalTrials.gov (aka CT.gov) provides a comprehensive search capability to enable the users to identify relevant trials by indication, compound, phase, year, and other criteria. It has a query expansion mechanism to expand search terms to semantic classes and synonyms to improve recall. Recent models enable the users to leverage the power of informational retrieval coupled with deep learning-based ranking \cite{rybinski2020clinical}. However, these tools do not provide a network or graph-based perspective of clinical trials.

More recently, applications of knowledge graphs and graph embedding are becoming popular due to the advantages of graph representation learning in capturing latent relationships, reducing dimensionality, learning contextual features, and improving machine learning prediction \cite{yue2020graph}.

Recent research in clinical trial representation learning has led to significant progress in learning low-dimensional features that are useful in search and prediction \cite{chen2021ctkg}. However, these methods only apply traditional knowledge graph embeddings into clinical trials research, but do not capture the user preference or custom knowledge into the representation learning process. Therefore, search and prediction based on these embeddings do not necessarily reflect the real intent of the user and thus compromise the utility.

We address how to incorporate custom knowledge into the embedding training process to improve the embedding quality. We use NLP relevancy search to model the user preference of trials and trial similarity. The relevancy search result is transformed into custom nodes and edges on the knowledge graph. We then develop custom2vec, an algorithmic framework for learning embedding representations for custom nodes in the network. The algorithm uses a joint learning to preserve the full network structure while improve the similarity among relevant nodes (trials) on the graph. It models the custom nodes and their links as a custom semantic subgraph. It then uses a joint random walk sampling strategy to generate random walks in both the full graph and the subgraph, to minimize the joint loss function.

We measure the quality by several means, including 1) checking the similarity score distribution among the custom nodes (relevant trials), 2) examining the impact of custom embedding (joint random walks) on the overall graph structure by checking the similarity scores among the native links of the graph, and 3) using the embeddings to predict/recommend similar trials by user preference.

We conduct a series of experiments consisting of 5 models and 2 datasets. The results show that custom2vec can pull user-defined custom nodes close to each other in the embedding space and can better predict the custom links using an unsupervised ranking based recommendation algorithm.

Our contributions include: (1) A novel subgraph approach to customizing knowledge-graph embeddings; (2) A new learning objective (loss function) that simultaneously optimizes for the customization criteria as well as for the baseline relationships; (3) An approach to using the standard search engine output for user-specified queries (i.e. user customization needs) as the training data for customized embeddings; (4) Experiments to show that the customized embeddings better represent similarity between relevant trials in the customization space without negatively effecting the rest of the trials; 5) An unsupervised link prediction technique to better recommend relevant trials reflecting the user preference.

\section{Related Work}

\subsection{Knowledge Graph in Clinical Research}
A knowledge graph is a powerful mechanism that can capture complex semantic relationships among elements of large-scale data, including data that is extracted from text documents.
Knowledge graph representation is an emerging area for clinical research. One trend is to construct knowledge graph using nodes and edges extracted from different fields of the clinical trial protocol \cite{chen2021ctkg}. The nodes can be structured entities as well as unstructured text. Additionally, named entities extracted from the text (e.g., the eligibility criteria) can be additional nodes \cite{du2021covid}.

Another trend is to learn graph representation for machine learning applications \cite{chen2021ctkg}. Previous work has shown the advantage of graph embeddings to learn low dimensional compact features for trial outcome prediction \cite{du2021covid}. But they do not necessarily preserve the semantic meaning of trial similarity from the user’s perspective.

In summary, current methods are based on static graphs and do not consider the user’s specific requirements. For example, there is no modeling of direct trial-trial relationships in the knowledge graph. The relationships among trials are indirect and hidden. Therefore, we need customization methods and experimental validation of embeddings in clinical research.

\subsection{Knowledge Graph Customization}

The traditional graph embedding method provides an effective way to understand the complex graph data. But the learning procedure is disconnected from the target applications. To address this challenge, several methods have been proposed to customize graph embedding. For example, \cite{hou2019customized} proposed Customized Graph Embedding (CGE) to randomly sample paths and re-weight them through a neural network model to reflect their importance to a specific application. \cite{werner2021retra} introduced the Recurrent Transformer (RETRA), a neural encoder with a feedback loop to incorporate situation-specific factors for training custom embeddings. \cite{reese2021kg} developed customized knowledge graph for COVID-19 and applied node2vec \cite{grover2016node2vec} to train embeddings. However, these studies did not explore subgraph-based approach in customizing knowledge graph embedding.

\section{Method}
\subsection{Modeling of user interests}

There are over 400k clinical trials in the CT.gov. Given a user query, there could be thousands of clinical trials returned by the CT.gov or other search engines. Most of the time, the users are only interested in a small subset of clinical trials most relevant to their research needs, for example, finding competitor trials from sponsors of similar size, or finding phase 3 trials for a specific indication. It is important to capture the users’ interests to learn representation of clinical trials.

We leverage the advanced search in CT.gov to retrieve clinical trials that represent user preferences. For example, a clinical team investigating a new cancer therapy by comparing it with a standard of care, e.g., docetaxel \cite{borghaei2015nivolumab}, may search trials that put docetaxel on the comparison arm. Alternatively, the team may search the PubMed or domain journals to curate relevant trial data involving docetaxel.

Note that the modeling of user-preferred relevant trials can be flexible. It can be extended to any types of search results or user manually vetted annotations. The user can either validate the result by a search engine or provide a list of trials by manual curation.

\subsection{Clinical trial graph construction}
Given a set of clinical trial protocols, we construct a knowledge graph using 6 entity types: Trial ID (clinical trial node), indication, intervention, phase, sponsor, and endpoint. All entities are extracted from the corresponding sections of the protocol. Trial ID and phase are standard values from the CT.gov. We standardize the indications using the MeSH ontology, and the interventions, sponsors, and endpoints using simple syntactic consolidation. These entities are modeled as nodes. The direct relationships between those nodes are modeled as links. For example, a trial node is linked to one phase node, one or multiple indication nodes, and one or more sponsor nodes, etc., according to the trial protocol document.

The custom data, obtained from a search engine output, or from a user-vetting process, is integrated into the graph as “additional semantic links”. For example, a list of phase 3 trials on PD-1 inhibitors \cite{sunshine2015pd} for a specific disease are often considered as trials similar to each other and therefore there will be direct links among them. We call the new graph semantically enriched graph.

\begin{figure}[h]
  \centering
  \includegraphics[width=\linewidth]{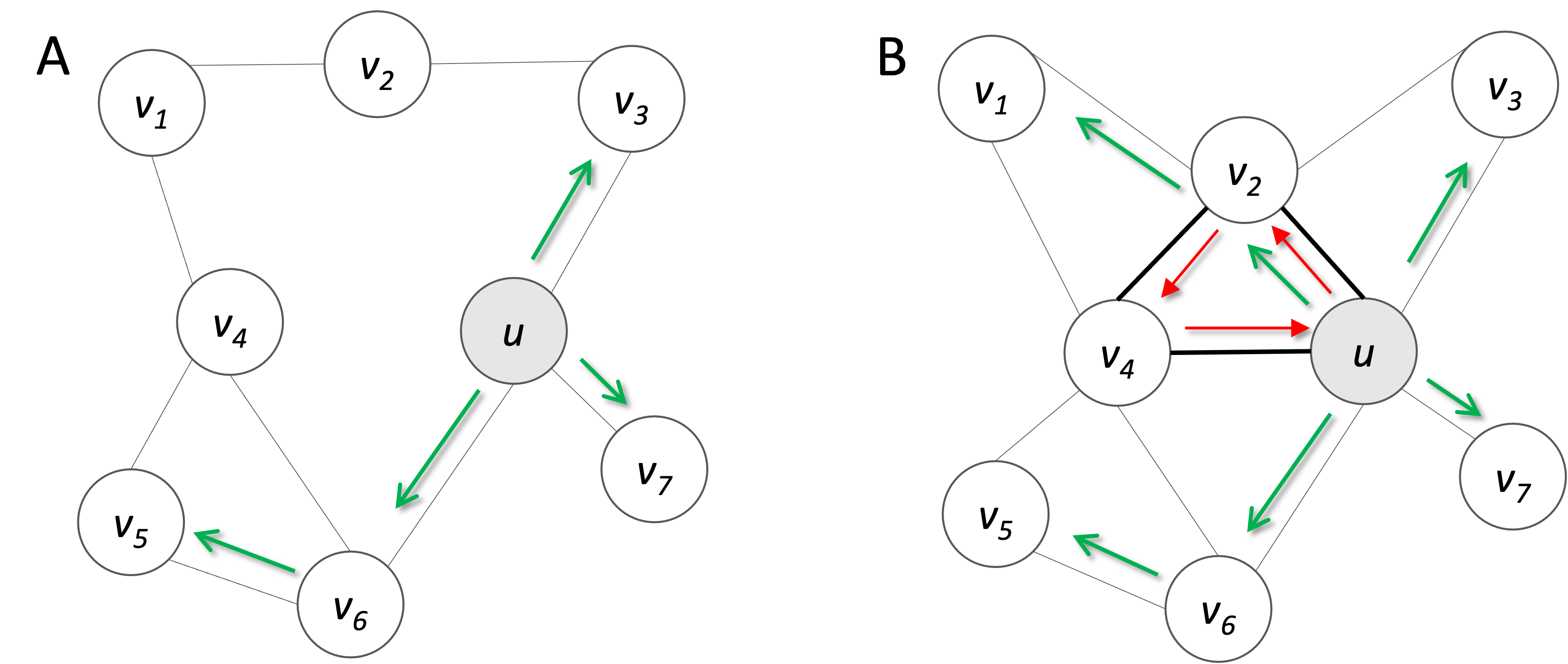}
  \caption{A) Conventional graph sampling with random walks (green path). B) Custom graph sampling with joint random walks (green path) in the full graph and random walks (red path) in the subgraph derived from user interests.}
  \label{fig:fig1}
\end{figure}

\subsection{custom2vec joint embedding learning}
We formulate custom node embedding (feature learning) in knowledge graph as a maximum likelihood optimization problem. Inspired by node2vec\cite{grover2016node2vec}, we propose joint random walks, including the full graph-based random walks, and the custom subgraph-based random walks. See Figure 1. The goal is to improve the likelihood of co-occurrence of nodes in the same random walk paths. For custom embedding, nodes in the subgraph will be closer to each in the embedding space, which reflects the user’s preferences. Nodes outside the subgraph but connected by full graph random walks will also be closer to each other to main the structure of the original neighborhood.

Specifically, we define nearby nodes $N_G (u)$ as neighborhood of node $u$ obtained by sampling of the full graph G, and $N_{SG} (u)$ as joint neighborhood of $u$ obtained by sampling the subgraph SG if $u$ is on SG. We can run shortest fixed length random walks starting from each node on G to collect $N_G (u)$ and $N_{SG} (u)$. Note if $u$ is not part of the subgraph, it does not have sub-graph based random walks, or $N_{SG} (u)$ is null.

The objective is to optimize embeddings to maximize the likelihood of random walk co-occurrences, we define the loss function as:

\begin{equation}
%   L = \sum_{i=0}^{\infty}x_i=\int_{0}^{\pi+2} f
    L = \sum_{u\in V}\sum_{v\in N_G(u)}-log(P(v|Z_u))+\sum_{u\in V'}\sum_{v'\in N_{SG}(u)}-log(P'(v'|Z_u))
\end{equation}

where $Z_u$ is the embedding of node $u$, $P(v|Z_u)$ is the probability of finding node $v$ in $N_G(u)$ given $Z_u$, and $P'(v'|Z_u )$ is the probability of finding node $v’$ in $N_{SG} (u)$ given $Z_u$.

We parameterize $P(v|Z_u)$ and $P(v'|Z_u)$ using softmax:

\begin{equation}
% P(v|Z_u )=exp(Z_u\cdot Z_v))/\sum_{n\in V}exp(Z_u\cdot Z_n)
P(v|Z_u )=\frac{exp(Z_u\cdot Z_v)}{\sum_{n\in V}exp(Z_u\cdot Z_n)}
\end{equation}

\begin{equation}
% P'(v'|Z_u )=exp(Z_u\cdot Z_{v'})/\sum_{n\in V'}exp(Z_u\cdot Z_n)
P'(v'|Z_u )=\frac{exp(Z_u\cdot Z_{v'})}{\sum_{n\in V'}exp(Z_u\cdot Z_n)}
\end{equation}

Put it together, we have:

\begin{equation}
\begin{split}
    % L = \sum_{u\in V}\sum_{v\in N_G(u)}-log(exp(Z_u\cdot Z_v))/\sum_{n\in V}exp(Z_u\cdot Z_n))\\
    % +\sum_{u\in V'}\sum_{v'\in N_{SG}(u)}-log(exp(Z_u\cdot Z_{v'})/\sum_{n\in V'}exp(Z_u\cdot Z_n))
    L = \sum_{u\in V}\sum_{v\in N_G(u)}-log(\frac{exp(Z_u\cdot Z_v)}{\sum_{n\in V}exp(Z_u\cdot Z_n)})\\
    +\sum_{u\in V'}\sum_{v'\in N_{SG}(u)}-log(\frac{exp(Z_u\cdot Z_{v'})}{\sum_{n\in V'}exp(Z_u\cdot Z_n)})
\end{split}
\end{equation}

To optimize custom embeddings, we need to find embeddings $Z_u$ that minimize $L$. As in conventional node2vec (Grover and Leskovec, 2016), we approximate the per-node function $\sum_{n\in V}exp(Z_u\cdot Z_n)$ and $\sum_{n\in V'}exp(Z_u\cdot Z_n)$ using negative sampling, for example, to sample k negative nodes proportional to degree to compute the loss function.

\subsection{Sampling strategy}
To optimize embeddings, we run joint random walks including node2vec type random walks that can tradeoff between local and global views of the full graph, as well as subgraph-based random walks that balance local and global views within the subgraph.

To focus on evaluating the effect of the joint graph and subgraph views, we use the default breadth-first sampling and depth-first sampling search strategies as in node2vec. We use the default return parameter $p$ $(p = 1)$ and in-out parameter $q$ $(q = 1)$. We also set the length of walks the same for both full graph walks and subgraph walks.

So now $N_G (u)$ and $N_{SG} (u)$ are the nodes visited by the guided walks. We then simulate $r$ random walks of length $l$ starting from each node $u$ and optimize the custom2vec objective using stochastic gradient descent.

\subsection{Ranking-based recommendation}
We developed a use case to recommend relevant trials on the knowledge graph through link prediction. The purpose is to leverage the embeddings learned from different models and evaluate their impact on the prediction performance.

We curated a dataset representing the user’s preference and translate them into links among relevant trials. So the data contains a list of trial-trial links representing the user preferred trial set. We split the data into a training set and a testing set. We combine the training set to the original graph to build a semantically enriched graph and apply standard node2vec to generate embeddings (node2vec enrich). We also apply custom2vec using different number of random walks to generate the embeddings. Then we use the embeddings to measure the similarity between trial nodes and rank the top N trial-trial links. We assume the top ranked trial-trial links are the recommended relevant trials (this will enable trial comparative study). We measure the prediction performance using $precision @ k$, by comparing the predicted links in the top $k$ result with the test links.

The link prediction is based on unsupervised similarity ranking method. In the future, we will test supervised learning for link prediction.

\section{Experiments}
\subsection{Case study: NSCLC clinical trial graph}
We used the non-small cell lung cancer (NSCLC) use case to evaluate algorithms. The raw data contained 5,725 trials by searching the CT.gov as of Dec 7, 2021. We constructed a knowledge graph using 6 entity types: Trial ID (clinical trial node), indication, intervention, phase, sponsor, and endpoint. There were 38,108 nodes and 75,509 edges in the original knowledge graph.

\subsection{Customization scenarios}

We considered two customization scenarios. The first was recommending immuno-oncology trials evaluating PD-1 inhibitors. By incorporating prior knowledge about PD-1 trials into the embedding training process, we demonstrated that our system was able to recommend relevant PD-1 trials on the knowledge graph.

The second was exploring similar trials that compare new therapies with a standard of care, e.g., docetaxel. We demonstrated that incorporating custom knowledge about docetaxel trials, the system is able to recommend more relevant trials involving docetaxel.

\subsection{Datasets}

\begin{table*}
  \caption{Data Sets}
  \label{tab:commands}
  \begin{tabular}{c|p{20mm}| p{20mm}|p{20mm}| p{20mm}|p{20mm}|p{20mm}}
%   \begin{tabular}{m{1cm}|m{1cm}| m{1cm} |m{1cm}| m{1cm}|m{1cm}|m{1cm}}  
    \toprule
    Dataset	&Enriched graph \newline nodes	&Enriched graph \newline edges	&train subgraph \newline nodes	&train subgraph \newline edges	&test subgraph \newline nodes	&test subgraph \newline edges\\    
    \midrule
    \texttt PD-1	&38108	&76785	&57	&1276	&57	&320\\    
    \texttt Docetaxel	&38108	&82853	&136	&7344	&136	&1836\\
    \bottomrule
  \end{tabular}
\end{table*}

We test our models on the following datasets:

\begin{itemize}

\item {\verb|PD-1|}: Using Phase 3 PD-1 trials from CT.gov search result as custom data. We construct a fully connected subgraph using the PD-1 trials as nodes. Then the subgraph is split into the train subgraph and test subgraph. The links of the train subgraph are added to the original graph to form the enriched full graph.

\item {\verb|Docetaxel|}: Using Phase 3 trials comparing new therapies with docetaxel from CT.gov search as custom data. We construct a fully connected subgraph using the docetaxel trials as nodes. Then the subgraph is split into the train subgraph and test subgraph. The links of the train subgraph are added to the original graph to form the enriched full graph.

\end{itemize}

The size of the graphs is shown in Table 1. The links in the train subgraph were added to the original graph to form the enriched graph. We trained embeddings on the enriched full graph using different models. The links in the test subgraph were used to measure the prediction performance of the trial recommendation using link prediction.

\subsection{Custom2vec Parameters setting}
We use the following parameters in custom2vec:

\begin{itemize}
\item {\verb|full graph parameters|}: $dimensions=20$, $walk length=16$, $num\_walks=100$
\item {\verb|subgraph parameters|}: $dimensions=20$, $walk length=16$, $num\_walks=\{100, 500, 1000\}$
\end{itemize}

We want to test the number of random walks in the subgraph, because intuitively the more walks, the more possible the relevant trials in the subgraph will co-occur in the random walks, which play a role to influence the loss function. We test num of walks as 100, 500, and 1000 respectively.

\subsection{Embedding models}
We tested 5 different models, see Table 2. Node2vec raw is a node2vec embedding of the original clinical trial graph. Node2vec enriched is a node2vec embedding of the enrich clinical trial graph. There are different versions of custom2vec models with varying numbers of random walks in the train subgraph.

\begin{table}[h]
  \caption{Embedding Models}
  \label{tab:commands}
  \begin{tabular}{ccl}
    \toprule
    ID &Model Name & Description\\
    \midrule
    \texttt 1& node2vec raw & Node2vec on raw graph \\
    \texttt 2& node2vec enriched & Node2vec on enriched graph\\
    \texttt 3& custom2vec 100& Custom2vec with 100 subgraph walks\\
    \texttt 4& custom2vec 500& Custom2vec with 500 subgraph walks\\
    \texttt 5& custom2vec 1000& Custom2vec with 1000 subgraph walks\\
    \bottomrule
  \end{tabular}
\end{table}

\section{Results and Discussion}

We conducted experiments on 2 datasets and 5 models, on MacBook Pro with 2.3 GHz 8-Core Intel Core i9. Here we report the results for customization scenario 1 using the ‘PD-1’ set, and customization scenario 2 using the ‘docetaxel’ set.

\subsection{Customization Scenario 1 using ‘PD-1’ set}
\subsubsection{Trial node similarity by embeddings}
We examined the distribution of cosine similarity among custom trials in the training set. Figure 2 shows the distribution of cosine similarity among custom trials in the training set. Node2vec raw generates a wide range of similarity from 0 to 1, which means that the user-preferred similarity among custom trials is not well captured (Figure 2A). custom2vec models improve the similarity by pulling relevant trials closer to each other, with custom2vec 1000 achieving the best performance, see Figure 2B. This shows that custom2vec can better preserve the user preferred similarity. The same distribution pattern is found in the test set.

\begin{figure*}[h]
  \centering
  \includegraphics[width=\linewidth]{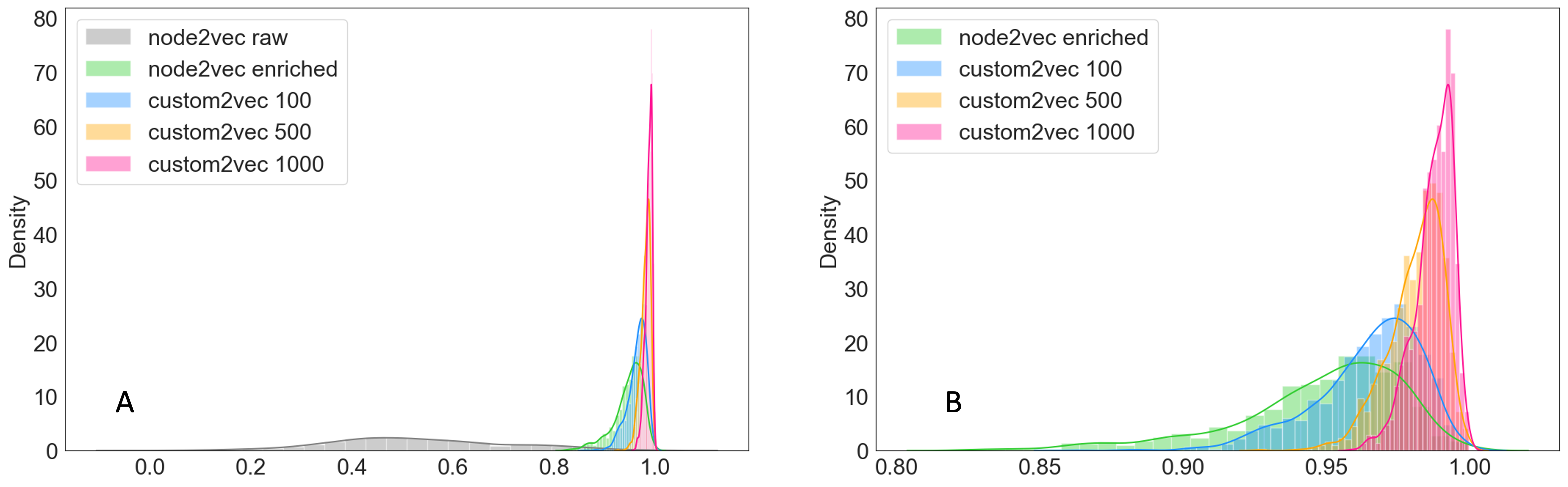}
  \caption{Distribution of the similarity between custom trial node pairs in the subgraph of the ‘PD-1’ dataset. A) Showing all embeddings. B) Showing embeddings on enriched graph only (no node2vec raw)}
\end{figure*}

\begin{figure*}[h]
  \centering
  \includegraphics[width=\linewidth]{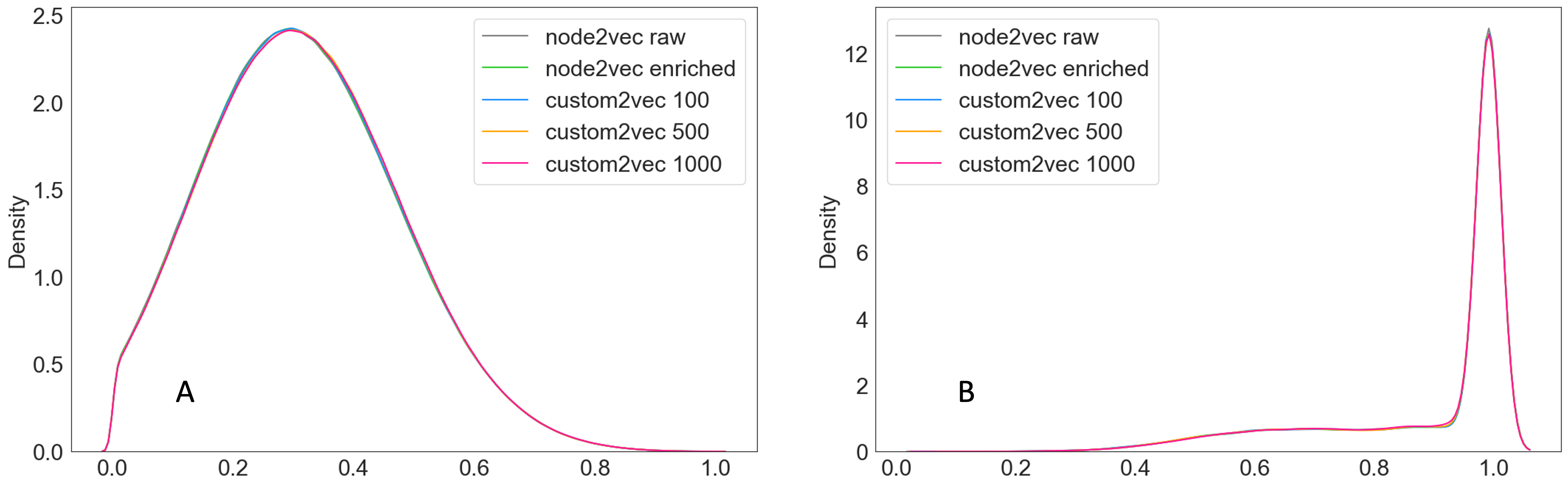}
  \caption{Distribution of the similarity between node pairs outside the subgraph of the ‘PD-1’ dataset. A) trial-trial similarity. B) trial-endpoint similarity}
\end{figure*}

\subsubsection{Analysis of native hidden links}

Some trials share a lot of common nodes, e.g., indication, endpoints, or test compounds, so they share similarity from the network perspective, which contrasts with node similarity from the user’s perspective. Since custom2vec also learns to maintain the full graph structure as in node2vec, those trial nodes are considered neighbors and should be similar in the embedding space.  For example, trials NCT04716933, and NCT03829319 are indirectly connected through common nodes (endpoints, indications, and sponsors). They have a cosine similarity of 0.99, meaning the native hidden links are preserved.

\begin{table}[h]
  \caption{Native link similarity of the ‘PD-1’ dataset}
  \label{tab:commands}
%   \begin{tabular}{ccl}
    % \begin{tabular}{m{5cm} | m{1cm}| m{1cm} |m{1cm}| m{1cm}}
    \begin{tabular}{c|c|c|c|c}
    % \hline
    % % \multicolumn{4}{|c|}{trial-trial, trial-endpoint} \\
    % % &trial-trial & &trial-endpoint\\
    % % \multicolumn{3}{| c |}{,tt,Begin of Table}\\
    % % Mode &  \multicolumn{2}{c}{Trial-Trial} & \multicolumn{2}{c}{Trial-Endpoint}\\
    % &  \multicolumn{2}{c|}{Trial-Trial} & \multicolumn{2}{c}{Trial-Endpoint}\\
    % % \hline
    \toprule
    % Model &trial-trial & &trial-endpoint\\
    % \midrule
    
    &  \multicolumn{2}{c|}{Native Trial-Trial} & \multicolumn{2}{c}{Native Trial-Endpoint}\\
    % \multirow{2}{*}{\textbf{Model}} &  \multicolumn{2}{c|}{Trial-Trial} & \multicolumn{2}{c}{Trial-Endpoint}\\
    
    \hline
    Model &Mean &Std &Mean &Std\\
    \midrule
    \texttt Node2vec raw &0.317 &0.158 &0.895 &0.164 \\
    \texttt Node2vec enriched &0.318 &0.158 &0.894 &0.165\\
    \texttt Custom2vec 100 &0.318 &0.158 &0.894 &0.165\\
    \texttt Custom2vec 500 &0.319 &0.158 &0.894 &0.164\\
    \texttt Custom2vec 1000 &0.319 &0.158 &0.894 &0.163\\
    \bottomrule
  \end{tabular}
\end{table}

\begin{table}[h]
  \caption{Native link similarity of the ‘docetaxel’ dataset}
  \label{tab:commands}
%   \begin{tabular}{ccl}
    % \begin{tabular}{m{5cm} | m{1cm}| m{1cm} |m{1cm}| m{1cm}}
    \begin{tabular}{c|c|c|c|c}
    % \hline
    % % \multicolumn{4}{|c|}{trial-trial, trial-endpoint} \\
    % % &trial-trial & &trial-endpoint\\
    % % \multicolumn{3}{| c |}{,tt,Begin of Table}\\
    % % Mode &  \multicolumn{2}{c}{Trial-Trial} & \multicolumn{2}{c}{Trial-Endpoint}\\
    % &  \multicolumn{2}{c|}{Trial-Trial} & \multicolumn{2}{c}{Trial-Endpoint}\\
    % % \hline
    \toprule
    % Model &trial-trial & &trial-endpoint\\
    % \midrule
    
    &  \multicolumn{2}{c|}{Native Trial-Trial} & \multicolumn{2}{c}{Native Trial-Endpoint}\\
    % \multirow{2}{*}{\textbf{Model}} &  \multicolumn{2}{c|}{Trial-Trial} & \multicolumn{2}{c}{Trial-Endpoint}\\
    
    \hline
    Model &Mean &Std &Mean &Std\\
    \midrule
    \texttt Node2vec raw &0.317 &0.158 &0.895 &0.164\\
    \texttt Node2vec enriched &0.318 &0.158 &0.893 &0.166\\
    \texttt Custom2vec 100 &0.319 &0.157 &0.892 &0.166\\
    \texttt Custom2vec 500 &0.320 &0.157 &0.892 &0.165\\
    \texttt Custom2vec 1000 &0.320 &0.157 &0.891 &0.166\\
    \bottomrule
  \end{tabular}
\end{table}

Figure 3A shows the distribution of cosine similarity among the native ‘indirectly’ connected trials. All the models share the same similarity pattern. Table 3 shows that the means and standard deviations of native trial-trial similarity distributions are similar for all embeddings. This means that custom2vec joint random walks do not significantly alter the original network structure in the embedding space.

\subsubsection{Analysis of native direct links}

We also examine how different embeddings impact the original direct links in the graph. Here we focus on examining the trial-endpoint links. For each trial-endpoint pair, we calculate its cosine similarity using different versions of embeddings (Figure 3B). As can be seen, all embeddings infer that those pairs are very similarity to each other, as the curve is highly skewed toward the upper limit of 1. This shows that custom2vec can also preserve the original direct links in the embedding space.

\subsubsection{Link prediction performance}

We measured the link prediction performance for 'PD-1' using the test subgraph, see Figure 6A. As can be seen, the node2vec raw model fails to predict user-expected trial-trial links. Node2vec enriched improves the prediction but is inferior to custom2vec predictions. Custom2vec 1000 achieves the best performance, with a precision of 0.10 for 10 predictions, a precision of 0.34 for 50 predictions, a precision of 0.38 for 100 predictions, and a precision of 0.514 for 1000 predictions. See Table 5.

\begin{table}[h]
  \caption{Link prediction using the test subgraph of the ‘PD-1’ dataset}
  \label{tab:commands}
    \begin{tabular}{c|c| c |c| c}
    \toprule
    Model &P@10 &P@50 &P@100 &P@1000\\
    \midrule
    \texttt Node2vec raw &0.00	&0.00	&0.00	&0.00 \\
    \texttt Node2vec enriched &0.00	&0.00	&0.01	&0.072\\
    \texttt Custom2vec 100 &0.00	&0.02	&0.02	&0.162\\
    \texttt Custom2vec 500 &0.10	&0.08	&0.11	&0.386\\
    \texttt Custom2vec 1000 &0.10	&0.34	&0.38	&0.514\\
    \bottomrule
  \end{tabular}
\end{table}

\subsection{Customization Scenario 2 using ‘docetaxel’ set}

\begin{figure*}[h]
  \centering
  \includegraphics[width=\linewidth]{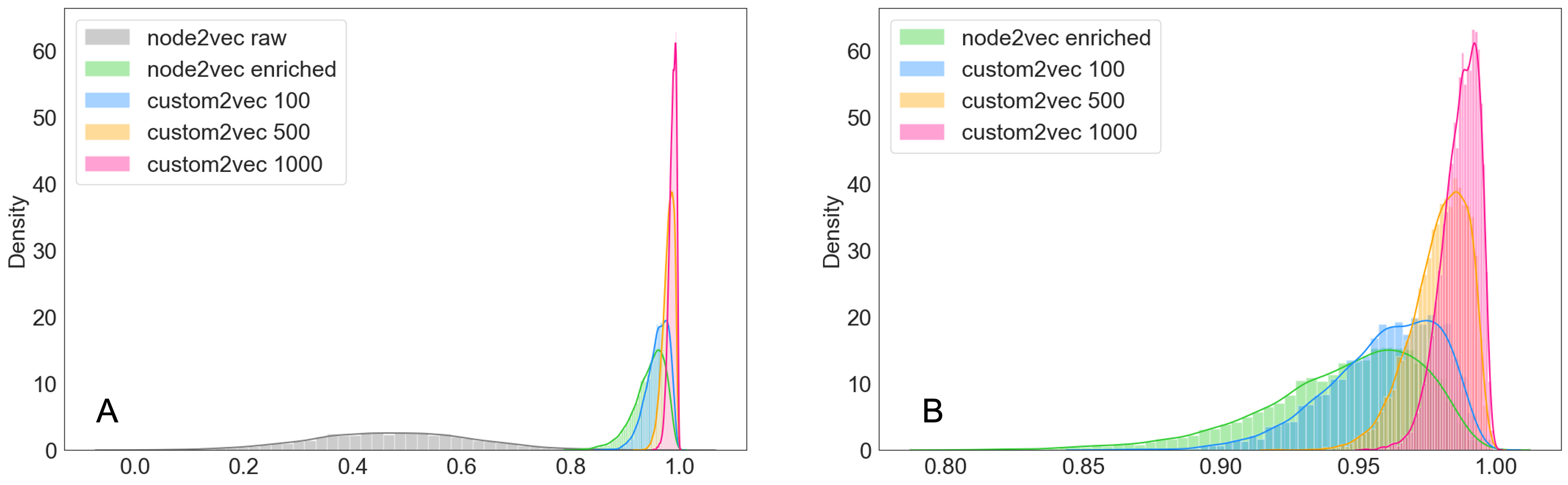}
  \caption{Distribution of the similarity between custom trial node pairs in the subgraph of the ‘docetaxel’ dataset. A) Showing all embeddings. B) Showing embeddings on enriched graph only (no node2vec raw)}
\end{figure*}

\begin{figure*}[h]
  \centering
  \includegraphics[width=\linewidth]{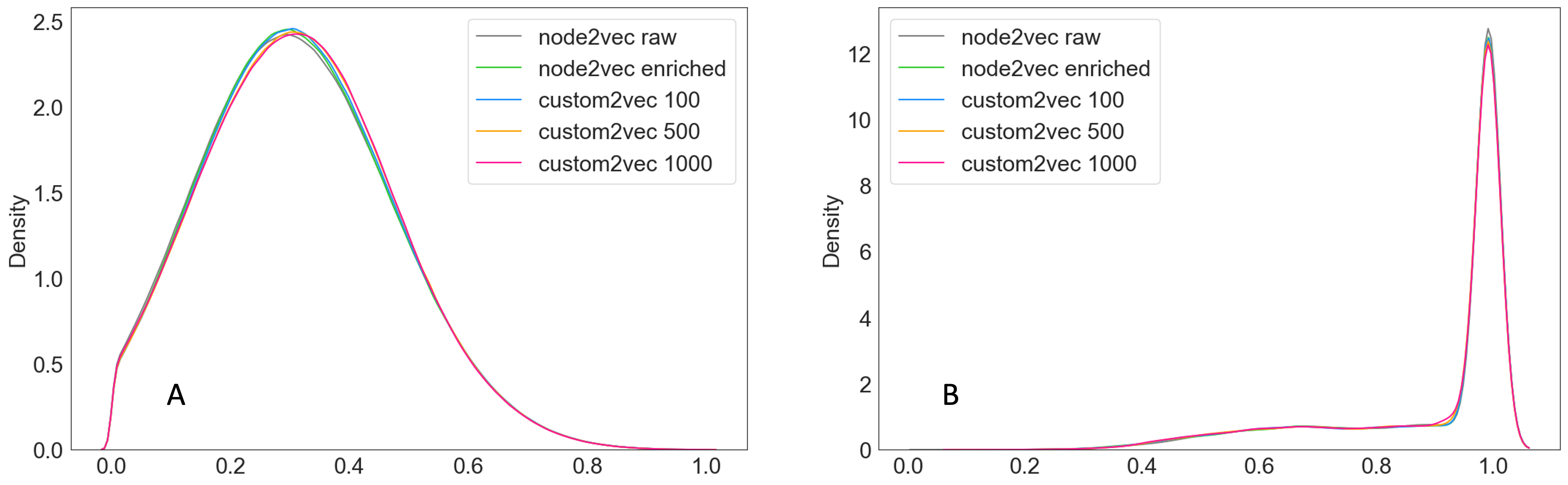}
  \caption{Distribution of the similarity between node pairs outside the subgraph of the ‘docetaxel’ dataset. A) trial-trial similarity. B) trial-endpoint similarity}
\end{figure*}

\subsubsection{Trial node similarity by embeddings}
Using the embeddings trained by different models, we can measure the similarity of custom trials (custom nodes). Figure 4 shows the distribution of cosine similarity among custom trials in the training set. Again node2vec raw generates a wide range of similarity from 0 to 1, which means that the user preferred similarity among custom trials is not well captured, see Figure 4A.

Therefore, node2vec on the raw clinical trial knowledge graph fails to capture the semantics of user preference. Node2vec on the enriched graph improves the similarity among custom trials with a higher average similarity score and a narrower standard deviation. While custom2vec models further improve the similarity by pulling relevant trials closer to each other, with custom2vec 1000 achieving the best performance, see Figure 4B. This shows that custom2vec can better preserve the user preferred similarity. The same distribution pattern is found in the test set.

\subsubsection{Analysis of native hidden trial-trial links}

Figure 5A shows the distribution of cosine similarity among the ‘indirectly’ connected trials. All the models share the same similarity pattern. Table 4 shows that the means and standard deviations of native trial-trial similarity distributions are similar for all embeddings. This means that custom2vec joint random walks do not significantly alter the original network structure in the embedding space.

\subsubsection{Analysis of native direct links}
We also examine how different embeddings impact the original direct links in the graph. Here we focus on examining the trial-endpoint links. For each trial-endpoint pair, we calculate its cosine similarity using different versions of embeddings (Figure 5B). As can be seen, all embeddings infer that those pairs are very similarity to each other, as the curve is highly skewed toward the upper limit of 1. This shows that custom2vec can also preserve the original direct links in the embedding space.

\begin{table}[h]
  \caption{Link prediction using the test subgraph of the ‘docetaxel’ dataset}
  \label{tab:commands}
%   \begin{tabular}{ccl}
    % \begin{tabular}{m{5cm} | m{1cm}| m{1cm} |m{1cm}| m{1cm}}
    \begin{tabular}{c|c| c |c| c}
    \toprule
    Model &P@10 &P@50 &P@100 &P@1000\\
    \midrule
    \texttt Node2vec raw &0.00 &0.00 &0.00 &0.00 \\
    \texttt Node2vec enriched &0.00 &0.00 &0.03 &0.507\\
    \texttt Custom2vec 100 &0.00 &0.02 &0.07 &0.615\\
    \texttt Custom2vec 500 &0.00 &0.28 &0.38 &0.765\\
    \texttt Custom2vec 1000 &0.30 &0.66 &0.74 &0.843\\
    \bottomrule
  \end{tabular}
\end{table}

% \begin{figure*}[h]
\begin{figure*}[ht]
  \centering
  \includegraphics[width=\linewidth]{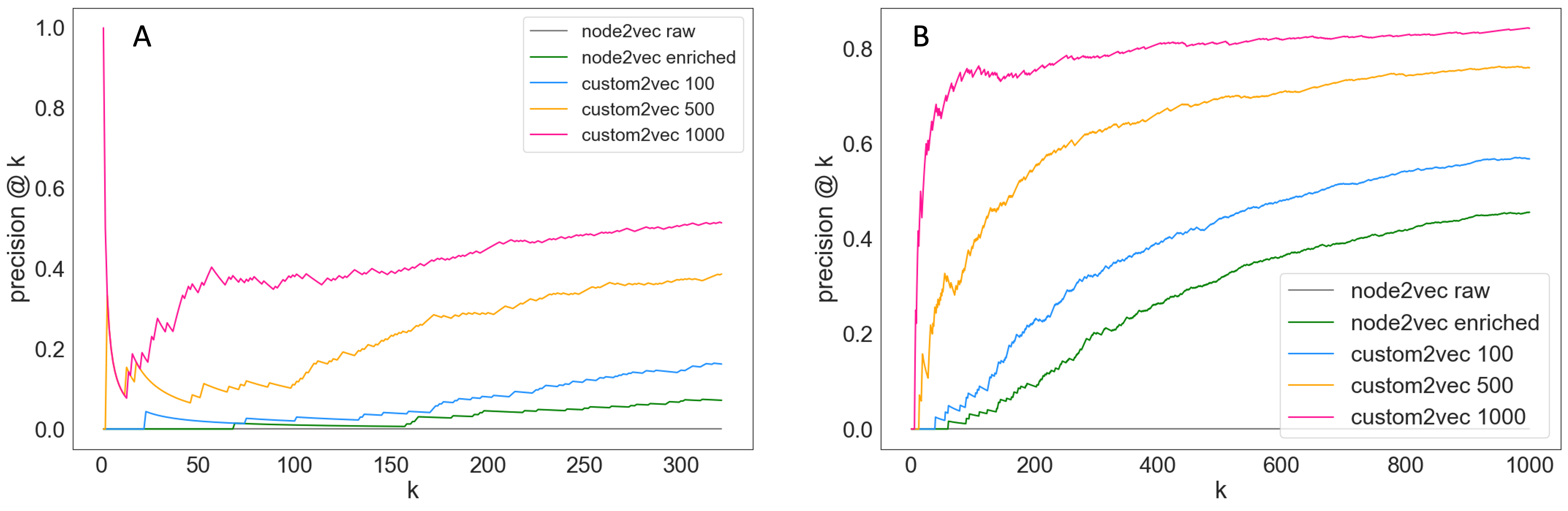}
  \caption{Link prediction performance using the test subgraph: A) the ‘PD-1' dataset. B) the 'docetaxel' dataset}
\end{figure*}

\subsubsection{Link prediction performance}
We measured the link prediction performance for 'docetaxel' using the test subgraph, see Figure 6B. Again, the node2vec raw model fails to predict user-expected trial-trial links. Node2vec enriched improves the prediction but is inferior to custom2vec predictions. Custom2vec 1000 achieves the best performance, with a precision of 0.30 for 10 predictions, a precision of 0.66 for 50 predictions, a precision of 0.74 for 100 predictions, and a precision of 0.843 for 1000 predictions. See Table 6.

\subsection{Summarization/discussion}
From the experiment results, we summarize the following observations:

\begin{itemize}

\item Custom2vec embeddings pull relevant trials closer to each other in both training and testing cases, thereby preserving the semantics of user preference. This allows the user to take advantage of the benefits of graph embeddings without losing the context of their specific research needs.

\item Custom2vec also preserves the original graph structure by maintaining the similarity of native hidden links (e.g., trial-trial) and native direct links (e.g., trial-endpoint) in the embedding space. This allows custom2vec embeddings to be used the same way as node2vec embeddings outside the custom subgraph in many down-stream applications.

\item The custom2vec embeddings allow for better prediction and recommendation of relevant trials through link prediction, because the user preferred links are highly ranked. For both 'PD-1' and 'docetaxel', custom2vec models systematically outperform node2vec and node2vec enriched in link prediction tasks.

\item The size of subgraph may impact the link prediction performance. In 'PD-1', there is a smaller subgraph with only 57 trials (nodes). The custom2vec 1000 model achieved the highest precision of 0.514 for 1000 predictions. While in the 'docetaxel' scenario, the subgraph is larger with 136 trials (nodes). And custom2vec 1000 achieved a precision of 0.843 for 1000 predictions. This suggests that more sufficient custom knowledge (subgraph with larger size) may contribute to more predictive custom embeddings. The detailed analysis of subgraph size and structure warrants future work.

\item Our use cases explored two exciting scenarios in lung cancer drug development to provide relevant insights about clinical trials. Our framework can be extended to more therapeutic areas and more data sources to facilitate the information needs of different clinical teams. This will accelerate clinical data package preparation and overall drug development.

\end{itemize}

\section{Conclusion}
We identified a clinical design scenario where traditional embeddings such as node2vec do not hold. We developed custom2vec with a composite objective to minimize the distance between user-preferred trials while maintain the original neighborhood structure. Detailed analysis shows that custom2vec can be used to customize knowledge graph embeddings and enable more accurate recommendations and predictions of relevant clinical trials.
Future work will include 1) exploring non-random walk based embedding methods, such as matrix factorization and graph neural networks; 2) studying more recommendation algorithms; and 3) conducting more experiments across therapeutics areas and phases.

% \usepackage[style=ACM-Reference-Format,backend=bibtex,sorting=none]{biblatex}.
% % \usepackage[style=ACM-Reference-Format,backend=bibtex,sorting=none]{biblatex}.
\bibliographystyle{unsrt}
\bibliography{references}

\appendix

\end{document}